\documentclass[letterpaper]{article} 
\usepackage{aaai25}  
\usepackage{times}  
\usepackage{helvet}  
\usepackage{courier}  
\usepackage[hyphens]{url}  
\usepackage{graphicx} 
\usepackage{natbib}  
\usepackage{caption} 
\usepackage{algorithm}
\usepackage{algorithmic}
\usepackage{latexsym}
\usepackage{amssymb}
\usepackage{amsmath}
\usepackage{amsthm}
\usepackage{booktabs}
\usepackage{enumitem}
\usepackage{graphicx}
\usepackage{color}
\usepackage{subfigure}
\usepackage{amsmath}

\usepackage{graphicx}
\usepackage{subcaption}
\usepackage{float}
\usepackage{newfloat}
\usepackage{listings}
\DeclareCaptionStyle{ruled}{labelfont=normalfont,labelsep=colon,strut=off} 
\floatstyle{ruled}
\newfloat{listing}{tb}{lst}{}
\floatname{listing}{Listing}
\title{Novelty-Guided Data Reuse for Efficient and Diversified Multi-Agent Reinforcement Learning}
\author{
    Yangkun Chen\equalcontrib,
    Kai Yang\equalcontrib,
    Jian Tao,
    Jiafei Lyu\thanks{Corresponding author.}
}
\affiliations{
    Shenzhen International Graduate School\\
    Tsinghua University\\

    \{chen-yk21, yk22, tj22, lvjf20\}@mails.tsinghua.edu.cn
}

\usepackage{bibentry}

\begin{document}
\maketitle

\begin{abstract}
Recently, deep Multi-Agent Reinforcement Learning (MARL) has demonstrated its potential to tackle complex cooperative tasks, pushing the boundaries of AI in collaborative environments. However, the efficiency of these systems is often compromised by inadequate sample utilization and a lack of diversity in learning strategies. To enhance MARL performance, we introduce a novel sample reuse approach that dynamically adjusts policy updates based on observation novelty. Specifically, we employ a Random Network Distillation (RND) network to gauge the novelty of each agent's current state, assigning additional sample update opportunities based on the uniqueness of the data.  We name our method Multi-Agent Novelty-GuidEd sample Reuse (MANGER). This method increases sample efficiency and promotes exploration and diverse agent behaviors. Our evaluations confirm substantial improvements in MARL effectiveness in complex cooperative scenarios such as Google Research Football and super-hard StarCraft II micromanagement tasks.
\end{abstract}

%
\begin{links}
    \link{Code}{https://github.com/kkane99/MANGER_code}
\end{links}

\begin{figure*}[t]
    \centering
    \subfigure[]{
    \begin{minipage}[t]{0.48\linewidth}
    \centering
    \includegraphics[height=0.465\linewidth]{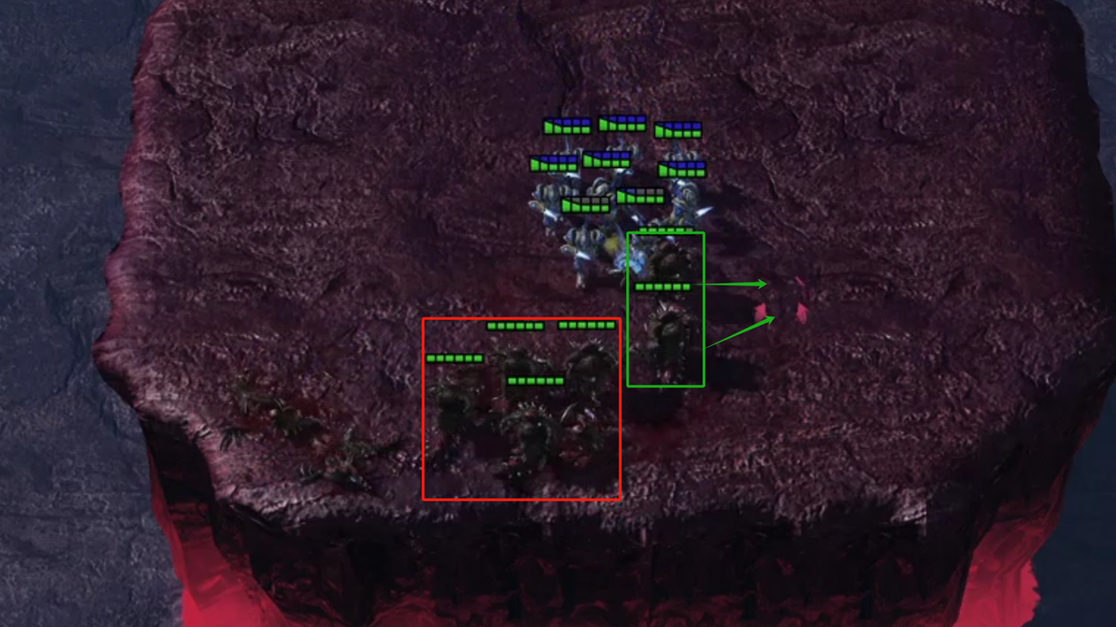}
    \end{minipage}
    }
    \subfigure[]{
    \begin{minipage}[t]{0.46\linewidth}
    \centering
    \includegraphics[height=0.483\linewidth]{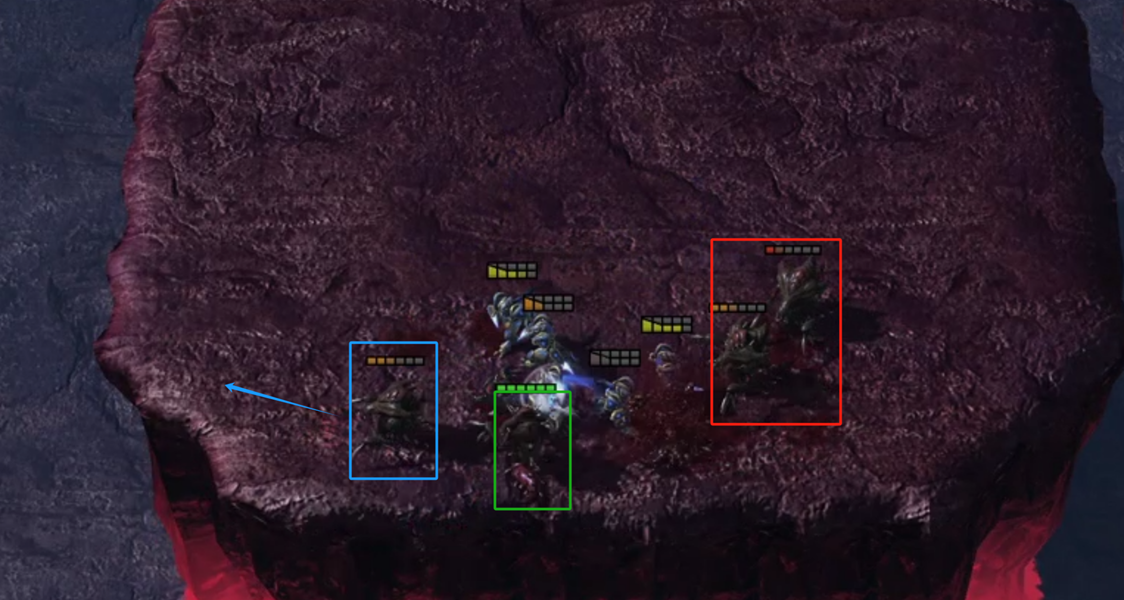}
    \end{minipage}
    }
    \caption{\textbf{Visualization of the environment}. In the SMAC image, the green box represents the tank role that actively absorbs damage and sacrifices itself to create an output environment for teammates. The red box represents the damage dealer role that activates attacks against enemies. The blue box represents the roaming role, similar to a guerrilla fighter, that can attract enemy aggro based on its own movement and lead some enemies away from the battlefield, preventing the enemies from focusing solely on the tank and causing it to be killed instantly.}
    \label{visualize}
\end{figure*}

\section{Introduction}

Reinforcement Learning (RL) \cite{sutton1998introduction} has emerged as a powerful paradigm in artificial intelligence, enabling agents to learn optimal behaviors through interactions with their environment. In recent years, Multi-Agent Reinforcement Learning (MARL) \cite{Canese2021MultiAgentRL} has attracted significant attention due to its applicability across various fields, including robotics \cite{Wang2023MultiRobotCS,Guo2022ExplainableAA,duan2024novelty}, autonomous vehicles \cite{Han2022StableAE,Zhang2022SpatialTemporalAwareSM,peng2021learning} and strategic games \cite{Jia2020FeverBA,Zhang2019HierarchicalRL,Vinyals2019GrandmasterLI,yang2023gtlma}.

MARL expands the principles of RL to scenarios involving multiple agents interacting with each other and the environment simultaneously. While traditional RL concentrates on a single agent that optimizes its behavior based on changes in the external environment, MARL necessitates the simultaneous control of multiple agents. The algorithm must account for the complexity of coordinating behaviors among these agents, making the task more challenging in terms of difficulty, training duration, and convergence \cite{yang2020qatten}.

In MARL, each interaction with the environment incurs higher costs and time, making strategies more challenging to learn and converge. For most MARL algorithms, data is stored in a buffer following an interaction with the environment. A batch of data is then sampled from the buffer for updates before continuing with the environment interaction. Due to the task's inherent complexity, immediately interacting with the environment after learning from a batch of data results in the strategy not fully utilizing the historical experience dataset. This leads to underfitting strategies during each interaction, resulting in no significant improvement in the quality of the sampled data and wasting interaction time and cost. To reduce the number of interactions with the environment and achieve desirable results, algorithms need to fully utilize historical data, train adequately, and then interact with the environment to obtain new, high-quality data.

Another critical aspect of MARL is the diversity among agents \cite{li2021celebrating,Bettini2024ControllingBD}. Unlike single-agent RL, which only needs to maximize its reward, some tasks require agents to play sacrificial roles, sacrificing individual gains for the best overall benefit. As seen in human societies, diverse skills, division of labor, and perspectives contribute to more effective collaboration. Similarly, in MARL, diverse agents can bring about a broader range of strategies and behaviors, ultimately leading to improved overall performance. Some recent work assigns specific roles to agents, executing certain actions to achieve diversity in agent strategies \cite{wang2020roma,Hu2022PolicyDV}. Others encourage exploration by maximizing the state entropy \cite{kim2023adaptive, tao2024multi} or other methods to diversify strategies among agents \cite{li2021celebrating, yang2024cmbe}. However, these methods require additional improvements to the algorithm's structure and do not perform well in complex environments.

To address these two key issues, we designed a method that improves the efficiency of sample utilization and improves diversity among agents. We have observed that the behavior exhibited by agents depends on the frequency of their policy updates. This phenomenon suggests that the update frequency of an agent's policy plays a crucial role in shaping their behavior and subsequent performance in cooperative tasks. By simply controlling the update frequency of each agent, diverse strategies can be achieved among agents. Additionally, repeated updates allow for the reuse of samples, thereby improving sample utilization.

To calculate the number of updates for each agent, the update frequency for each agent is tailored based on the novelty of their current observations. Specifically, we use a measure of state diversity to gauge the novelty of observations encountered by each agent. Novel or rare states trigger more frequent policy updates for that agent, while common or familiar states result in fewer updates. This adaptive update schedule serves three main purposes. First, it enhances sample efficiency by allowing agents to reuse data multiple times, effectively extracting more information from the historically visited dataset. Second, we posit that the variations in the overall value function are primarily driven by the value functions of agents encountering novel states. Therefore, additional updates are necessary to these agents to ensure a more thorough fitting. Lastly, the diversity in the number of update steps for each agent allows agents to have different behavioral strategies, promoting diversity in strategies among different agents, and allowing each agent to learn specific skills and division of labor for better cooperation.

By improving sample utilization and promoting diverse behaviors among agents, our method aims to enhance their cooperative abilities and increase the success rate of tasks that require coordinated action. Through empirical evaluation and theoretical analysis, we have demonstrated the effectiveness of our method in improving the efficiency and performance of MARL systems across various domains.

\begin{figure*}
    \centering
    \includegraphics[width=0.7\linewidth]{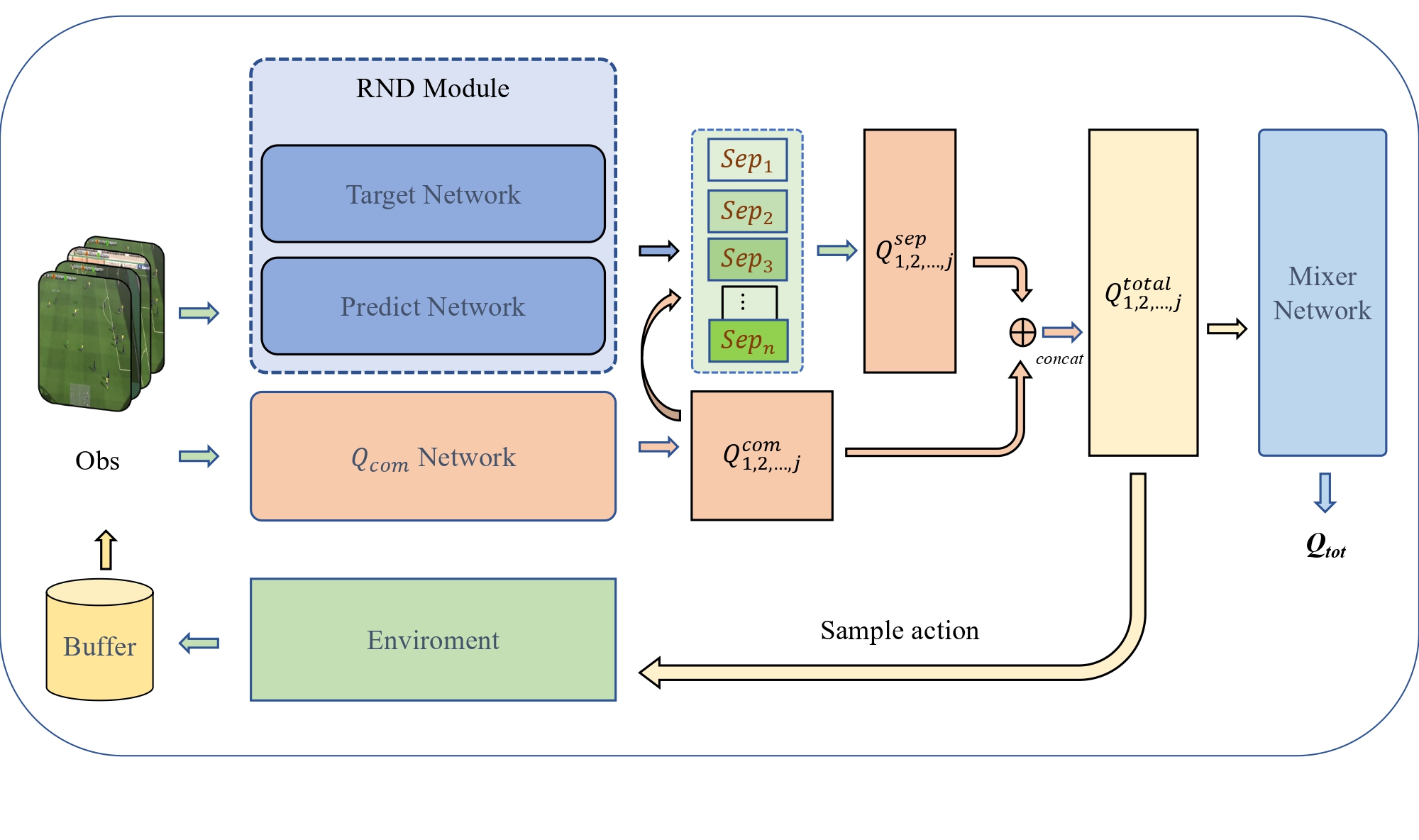}
    \caption{\textbf{Overview of the MANGER framework}. We employ the RND network to assess the novelty of each agent's observations, thereby enabling differentiated updates among agents. Furthermore, we ensure that each additional update does not interfere with the agents by decomposing the network.}
    \label{overview}
\end{figure*}

\section{Related Work}
\textbf{Multi-agent RL.} The MARL field offers diverse approaches to tackle challenges like non-smoothness, efficient communication, and the balance between suboptimality and decentralization in multi-agent systems. A widely used architecture is Centralized Training with Decentralized Execution (CTDE), where agents share historical data to enhance learning. CTDE assumes a central controller to process information from all agents, addressing the non-smoothness problem. Notable methods within this paradigm include MADDPG \cite{lowe2017multi}, MAPPO \cite{yu2022surprising}, VDN \cite{sunehag2017value}, QMIX \cite{rashid2020monotonic}, MAVEN \cite{mahajan2019maven}, and QPLEX \cite{wang2020qplex}. QMIX ensures monotonicity between global and individual Q-functions, while QPLEX relaxes the Individual-Global-Max (IGM) constraint with a duplex dueling network.

Research also focuses on extending MARL to handle high-dimensional, continuous state spaces via function approximators. Coordination graphs \cite{guestrin2001multiagent} factor large MDPs, enabling inter-agent communication through message passing, a concept supported by other studies \cite{lazaridou2020emergent, liu2020when2com}. To address the suboptimality-decentralization trade-off, methods like QTRAN \cite{son2019qtran} introduce relaxed penalties, while QAtten \cite{yang2020qatten} employs a multi-head attention-based Q-value mixing network. Finally, approaches like MAVEN \cite{mahajan2019maven} seek to encourage diverse behaviors through intrinsic rewards or hierarchical RL, maximizing mutual information between states and actions to foster agent diversity.

\noindent\textbf{Sample Efficiency.} In recent years, researchers have proposed various approaches to enhance sample efficiency in single or multi agent systems \cite{lyu2022efficient,lyu2023value,lyu2024cross,li2022prag,yan2024enhancing,gogineni2023accmer}. The Randomized Ensemble Double Q-learning (REDQ) algorithm \cite{chen2021randomized} demonstrates that employing a larger Update-To-Data (UTD) ratio can significantly boost sample efficiency, yielding substantial performance improvements compared to model-based reinforcement learning algorithms. To mitigate potential overfitting issues, REDQ utilizes an ensemble of Q-networks to reduce estimation errors in Q-values. Similarly, the Adaptive Value-Targeted Learning (AVTD) algorithm \cite{oh2021adaptive} employs a validation set to estimate the fitting error of the current network, adopting a more regularized approach with lower TD error to address overfitting problems associated with sample reuse. However, the Sample Multiple Reuse (SMR) algorithm \cite{lyu2024off} reveals that performing multiple repeated updates without an excessively large UTD value does not necessarily result in overfitting.

\noindent\textbf{Agent Diversity.} In MARL, agent diversity is crucial for enhancing system robustness, adaptability, and performance. This diversity, referring to variations in agents' strategies, objectives, or learning processes, significantly influences multi-agent interactions. Recent research highlights the importance of fostering diversity to better address complex environments. One effective method to promote agent diversity is through role specialization, where agents adopt specialized behaviors to cover a wider range of strategies. Notable examples include RODE \cite{wang2020rode}, which dynamically identifies roles based on state and action history, allowing agents to adapt their strategies to the environment's evolving needs. This method improves performance in cooperative tasks by facilitating scalable adaptation. ROMA \cite{wang2020roma} also enhances diversity by using a latent variable model to infer roles and a mutual information objective to promote diversification, resulting in more effective cooperation. ASN \cite{wang2019action} employs a behavior-based semantic neural network to calculate action semantics and achieve diversity by differentiating internal and external actions. The value of agent diversity is further emphasized in studies on emergent behaviors in unsupervised settings, where diverse behaviors can arise from simple reward structures in complex environments \cite{haber2018learning}. This emergent diversity improves adaptability and system resilience. In competitive settings, training agents with diverse policies through population-based training leads to more robust strategies that generalize across various opponents \cite{baker2019emergent}. Additionally, DIYAN \cite{eysenbach2018diversity} explores training agents to maximize behavioral diversity without explicit rewards, fostering a broad range of skills adaptable to various tasks.

In conclusion, agent diversity is essential for developing flexible and robust MARL systems. Our work aims to enhance adaptability and performance in dynamic environments by emphasizing the importance of diversity and effective discovery methods.

\section{Preliminaries}
\textbf{Dec-POMDP}: Our approach frames a fully cooperative multi-agent task within the context of a decentralized partially observable Markov decision process (Dec-POMDP) \cite{oliehoek2016concise}, described by the tuple $M = \langle N, S, A, P, r, Z, O, \gamma \rangle$. In this formulation, $N$ signifies a finite set of agents, $s_t \in S$ denotes the global state of the environment and $\gamma \in [0, 1)$ serves as the discount factor. At each time step, individual agents $j \in N$ perceive their own observations $o \in O$ and subsequently determine actions $a_k \in A$ based on the current global state $s_t$, where $k \in {1,2,...,|\mathcal{A}|}$ delineates the action space's capacity. These individual actions coalesce into a joint action vector $a_t$ at time step $t$. The resultant joint reward, $r(s_t,a_t)$, triggers a transition in the environment as dictated by the transition function $P(s' | s, a_t)$. Subsequently, the joint policy $\pi$ generates a joint action-value function: $Q_{\text{tot}}^{\pi}(s,a) = \mathbb{E}{s_{t:\infty},a_{t:\infty}}[G_t|s_t=s,a_t=a,\pi]$, where $G_t = \sum_{t}^{\infty} \gamma^t r_{t+1}$ represents the expected discounted return. This comprehensive representation encapsulates the collaborative decision-making dynamics inherent in multi-agent systems.

\noindent\textbf{Novelty} We use the frequency of state visits as a measure of novelty. Many articles have analyzed this aspect of state novelty measurement, such as ICM \cite{pathak2017curiosity}, RND \cite{burda2018exploration}, CFN \cite{lobel2023flipping}, RCMP \cite{da2020uncertainty} and DRND \cite{yang2024exploration}. In this paper, we employ the RND algorithm, a curiosity-driven mechanism, as the evaluation criterion for the novelty of current states or observations. Specifically, we have an untrained target network $f_\text{target}(o)$ and a network to be trained $f_\text{predictor}(o)$. During training, when the observation o has been seen many times, the mean squared error (MSE) loss between their predicted values tends to be small, and vice versa. Therefore, the higher the MSE loss, the higher the novelty of the observation. In this paper, we use this MSE loss to be the indicator of the novelty of each agent's observation.

\section{Method}

In multi-agent reinforcement learning, almost all methods interact with the environment after updating parameters once with the data, without fully utilizing the samples. Furthermore, commonly used algorithms like QMIX, MAPPO, and MADDPG employ parameter-sharing techniques to reduce training costs and achieve some training effectiveness quickly, but they overlook the diversity of roles among agents, hindering the learning of more complex and better cooperative strategies. In this chapter, we propose MANGER based on the QMIX algorithm, which can determine the efficiency of sample utilization based on the novelty of each agent's state, thereby allowing for different performances among different agents and improving sample utilization rates.

\subsection{Assessing the Novelty of Observations}
We employ RND to evaluate the novelty of agents' observations. Initially, we initialize a trainable predictor network $f_\text{predictor}$ and a random, fixed target network $f_\text{target}$. When agents interact with the environment and receive observations, the novelty of observation $o_i$ for agent $i$ is calculated as follows:
\begin{equation}\label{eq1}
N(o_i) = \|f_\text{target}(o_i)-f_\text{predictor}(o_i)\|^2.
\end{equation}
The RND algorithm was originally designed to address the challenge of sparse rewards in environments by measuring the novelty of states and using intrinsic rewards to encourage agent exploration. However, this paper does not focus on exploration or sparse reward problems. RND is used solely as a metric to evaluate the novelty of observations.

Note that we utilize only a single total predictor and one target network instead of training individual predictors and preparing specialized target networks for each agent. This decision stems from the fact that when an observation $o_i$ has been encountered multiple times, regardless of whether it was observed by other agents, we do not want to encourage the agent to consider it novel and reuse the data excessively. If an observation has been predominantly visited by agent $i$, reducing its novelty is reasonable as it indicates the observation has been visited frequently. Conversely, if the observation has been primarily visited by other agents, maintaining a low novelty prevents agent $i$ from mimicking the behavior of those agents, thus enhancing the diversity among agents. Subsequently, after computing the novelty of each observation across agents, we leverage this information to determine which data can be reused efficiently and how many times it is appropriate to reuse it.

\subsection{Using Data Efficiently}
Interacting with multi-agent environments is a time-consuming process, making the extra utilization of samples particularly important. From a holistic perspective, we enhance sample utilization rates by updating the overall framework with each data point twice. Additionally, considering each agent individually, we determine the novelty of data from each agent based on the aforementioned calculation of observation novelty, thereby deciding whether to reuse it. For data that has already appeared repeatedly, the network's estimates are highly accurate, and thus further updating is unnecessary. Conversely, when agents encounter previously unseen observations, such as scoring a goal in a soccer environment or enemy units being destroyed in a StarCraft environment, since these states have rarely occurred before, we need to artificially update such data more frequently to ensure more accurate estimates under these observations. Additionally, leveraging these data more can encourage agents to explore these previously unseen states more, thereby enabling them to better learn challenging cooperative strategies and find globally optimal solutions. The criterion for performing additional updates is as follows: first, the novelty values $N_\text{total}={N_1,N_2,...N_M}$ within a batch are normalized, and then, based on the relationship between the normalized values and their variance, the number of additional updates is determined. The formula for the number of times samples are additionally updated is as follows:
\begin{equation}\label{eq2}
T_i = \text{int}\left(\alpha \frac{(N_i-\bar N_\text{total})}{\text{Var}(N_\text{total})}\right).
\end{equation}

Here, $\alpha$ is a coefficient controlling the rates of additional data reuse, $\bar N_\text{total}$ represents the mean value of $N_\text{total}$, and $\text{Var}(N_\text{total})$ denotes the standard deviation of $N_\text{total}$. When $T_i$ is less than 1, we do not perform extra updates on the network of agent $i$. In this study, we set $\alpha=2$ and observe that the mean number of extra updates is less than 0.5, which does not significantly increase the training time.

\begin{figure*}[t]
    \centering
    \includegraphics[width=0.25\linewidth]{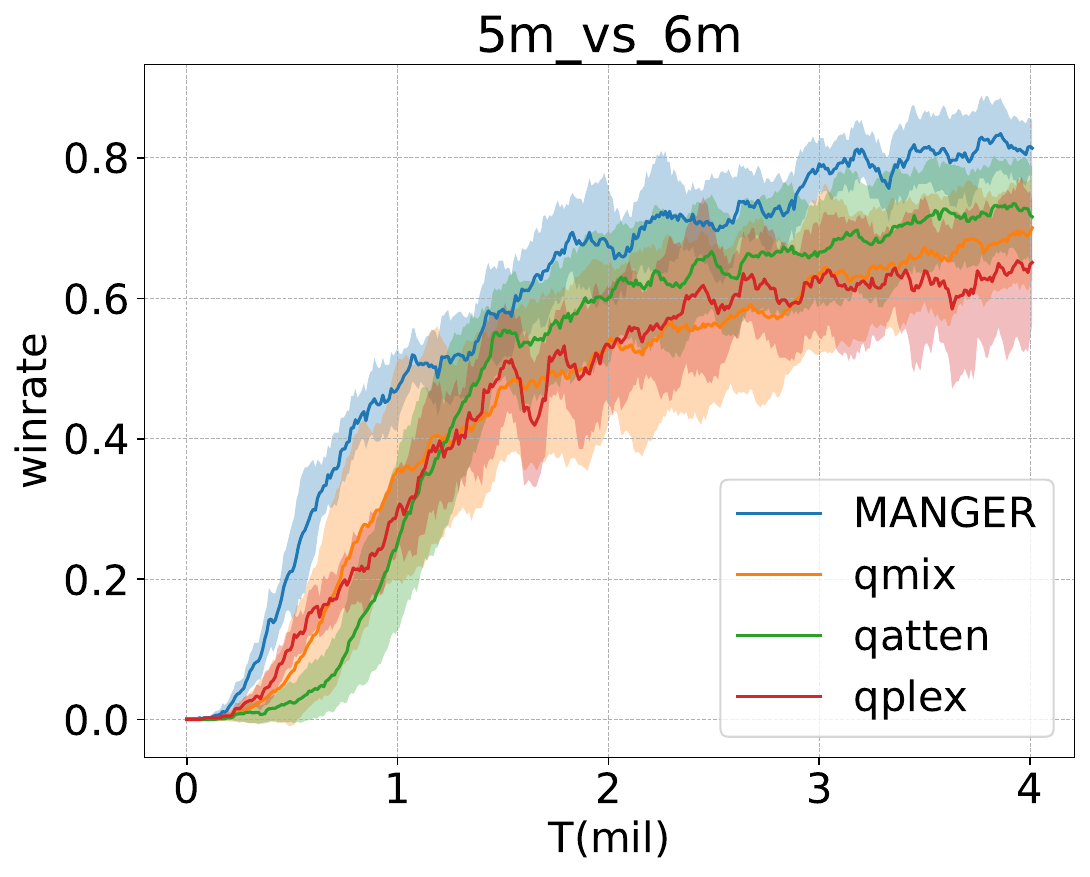 }
    \includegraphics[width=0.24\linewidth]{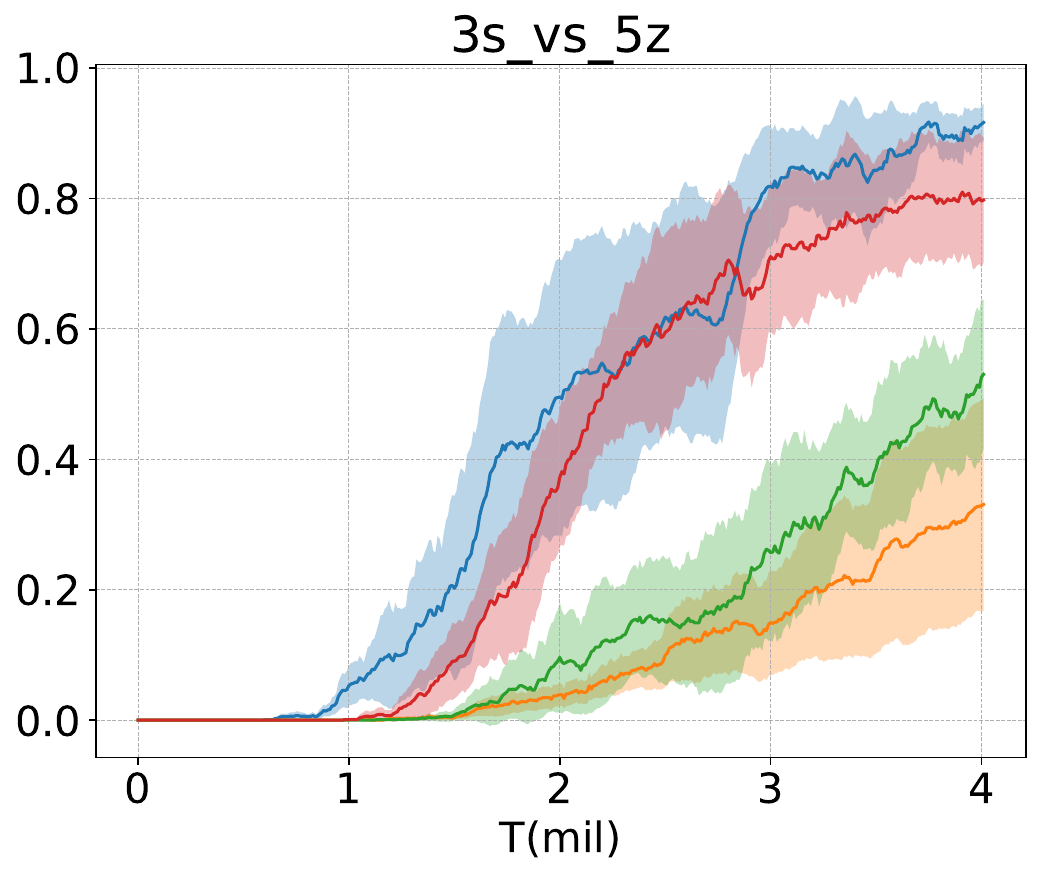}
    \includegraphics[width=0.24\linewidth]{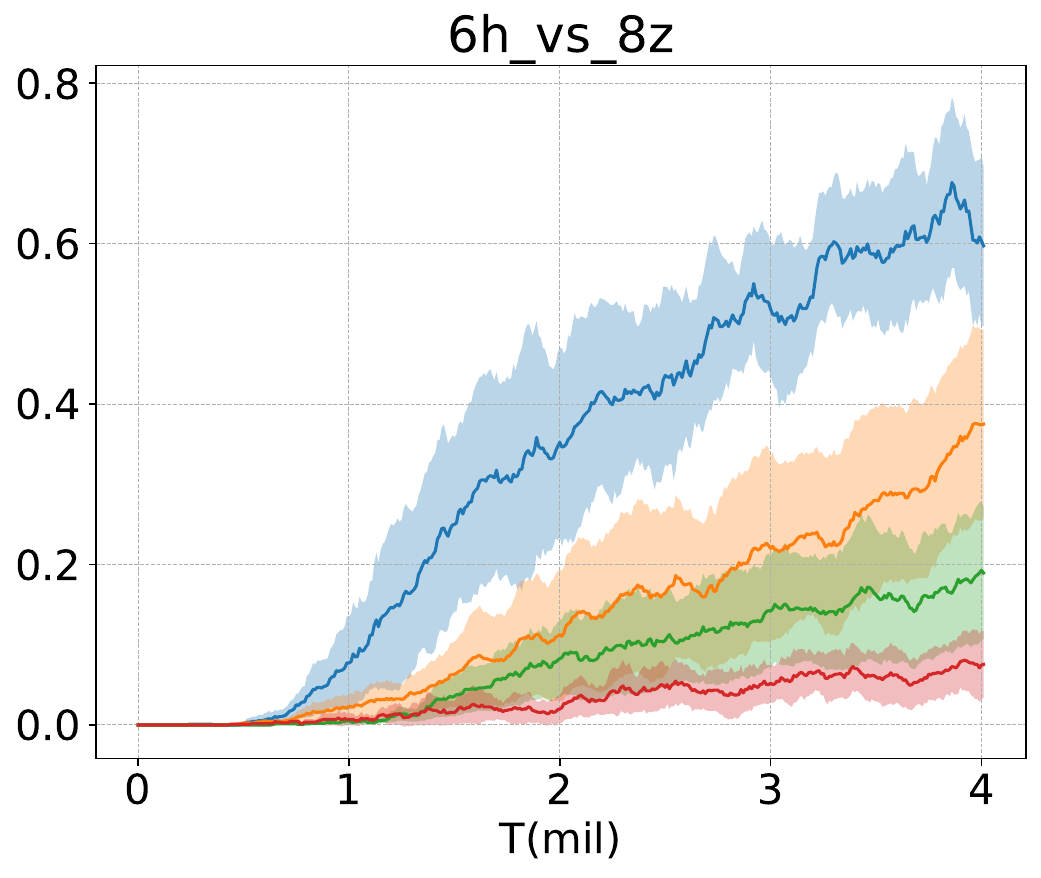}
    \includegraphics[width=0.24\linewidth]{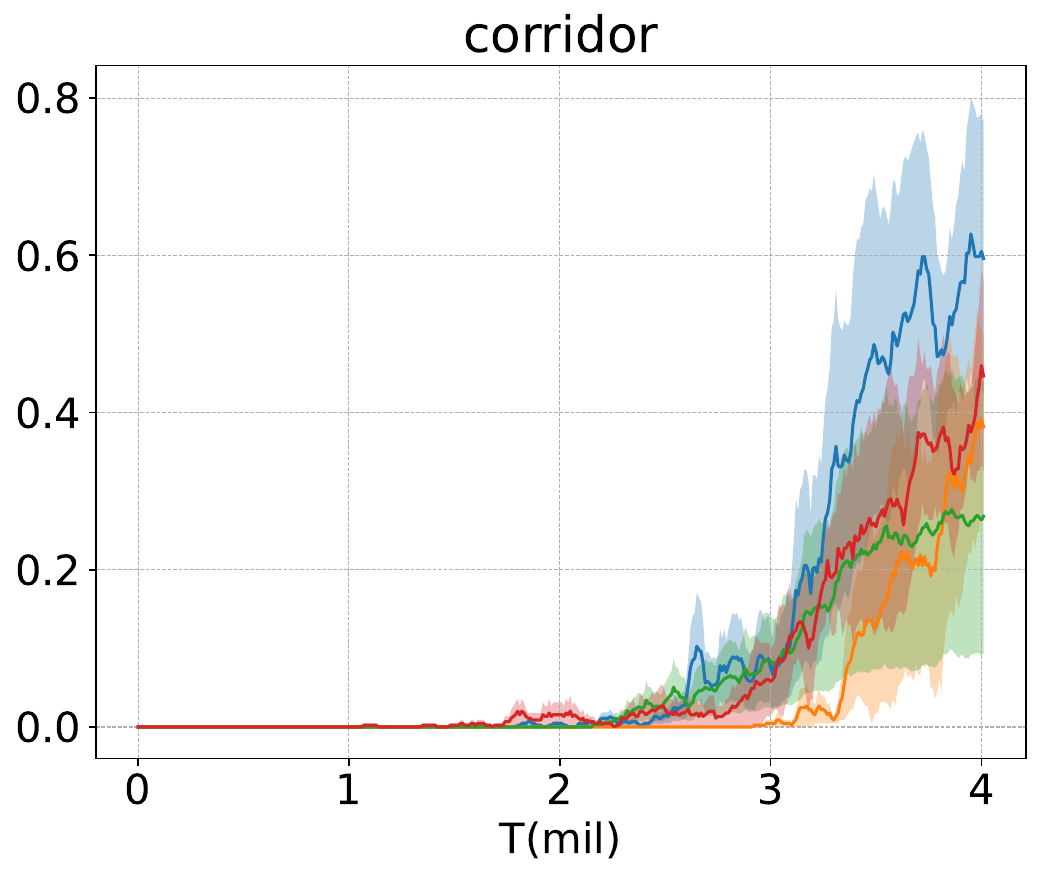}
    \caption{\textbf{Experimental results on SMAC}. All curves are averaged over 5 independent runs.}
    \label{fig:exp_on_smac}
\end{figure*}
\begin{figure*}[t]
    \centering
    \includegraphics[width=0.30\linewidth,height=0.252\linewidth]{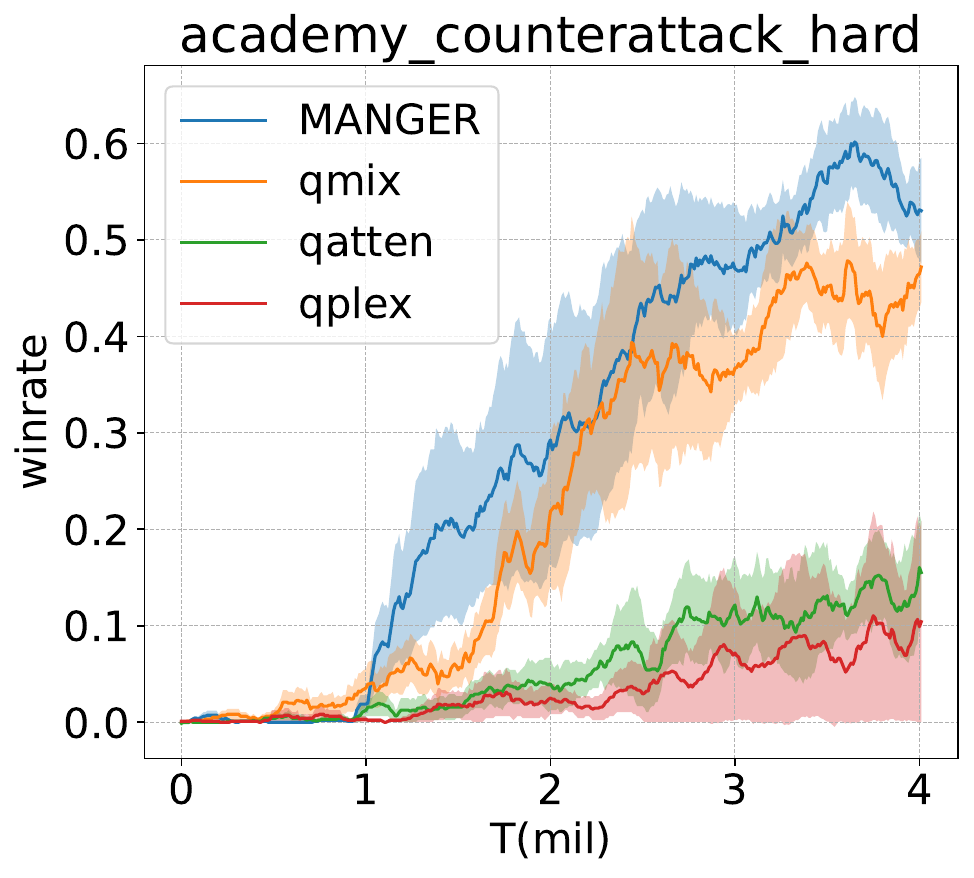 }
    \includegraphics[width=0.28\linewidth,height=0.25\linewidth]{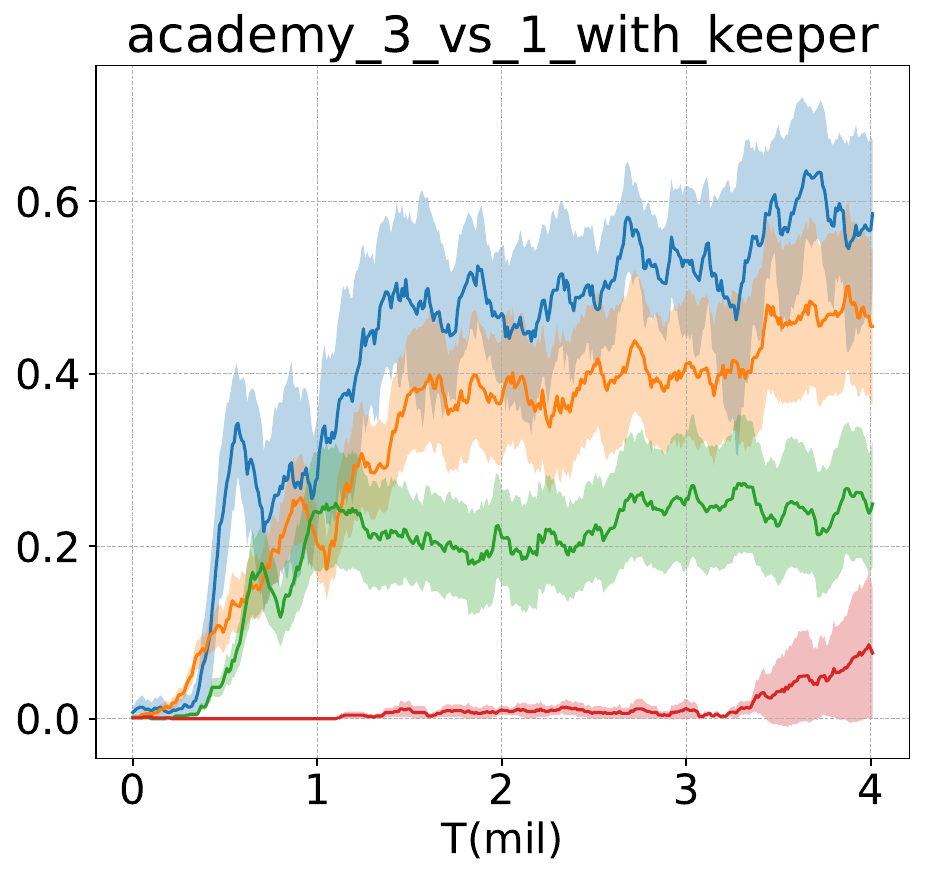}
    \includegraphics[width=0.28\linewidth,height=0.25\linewidth]{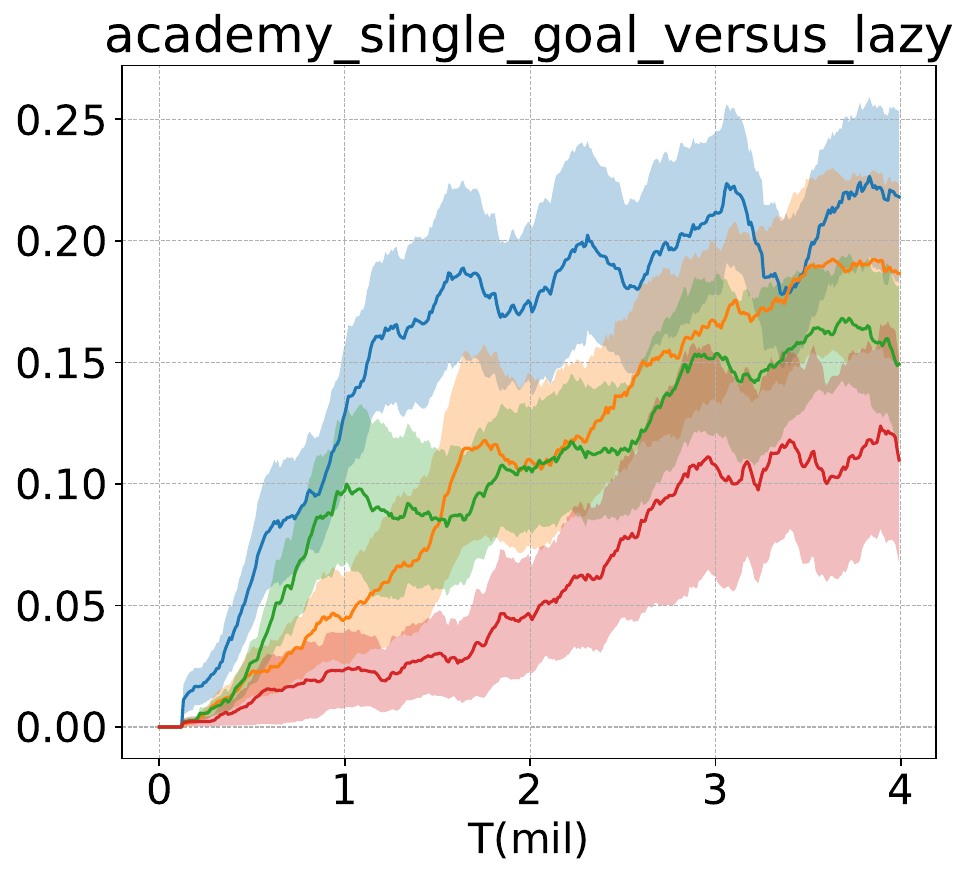}
    \caption{\textbf{Experiments on GRF environments}. All curves are averaged over 5 independent runs.}
    \label{fig:exp_on_GRF}
\end{figure*}

\subsection{Diversify the Agents by Separating the Critic Network}

\begin{figure*}[t]
    \centering
    \subfigure[]{
    \begin{minipage}[t]{0.8\linewidth}
    \centering
    \includegraphics[width=0.9\linewidth]{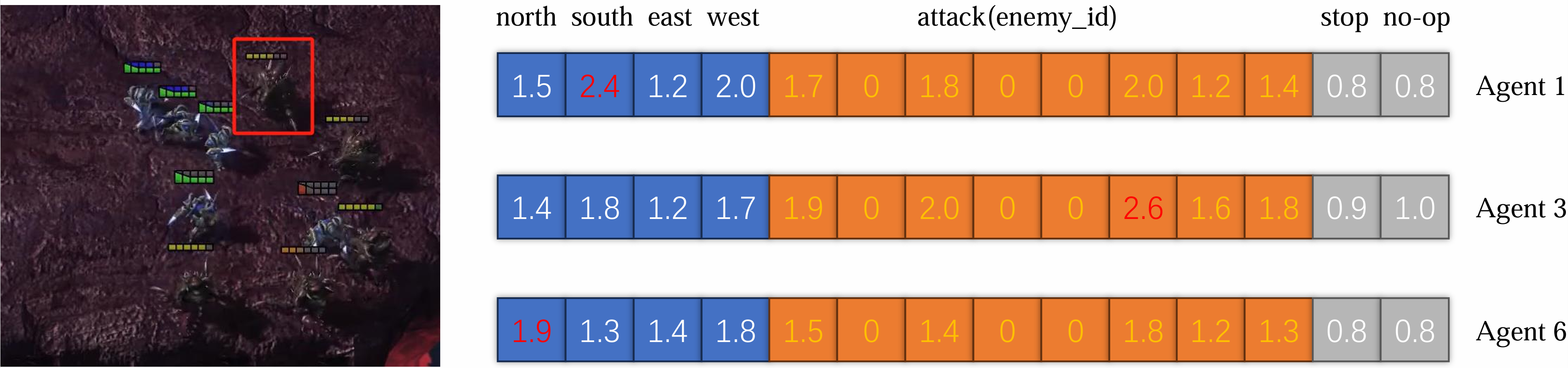}
    \end{minipage}
    }
    \subfigure[]{
    \begin{minipage}[t]{0.48\linewidth}
    \centering
    \includegraphics[height=0.56\linewidth]{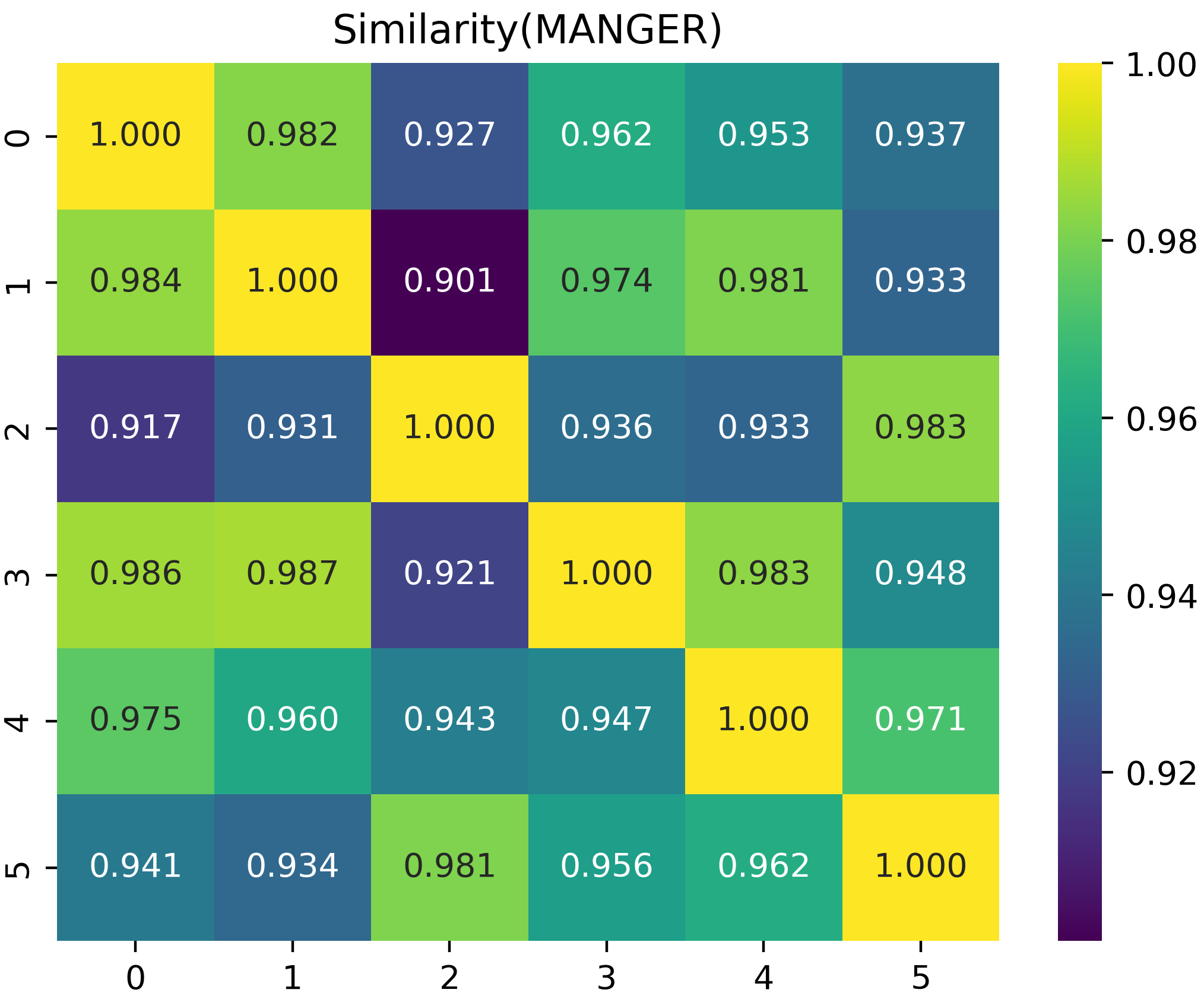}
    \end{minipage}
    }
    \subfigure[]{
    \begin{minipage}[t]{0.46\linewidth}
    \centering
    \includegraphics[height=0.56\linewidth]{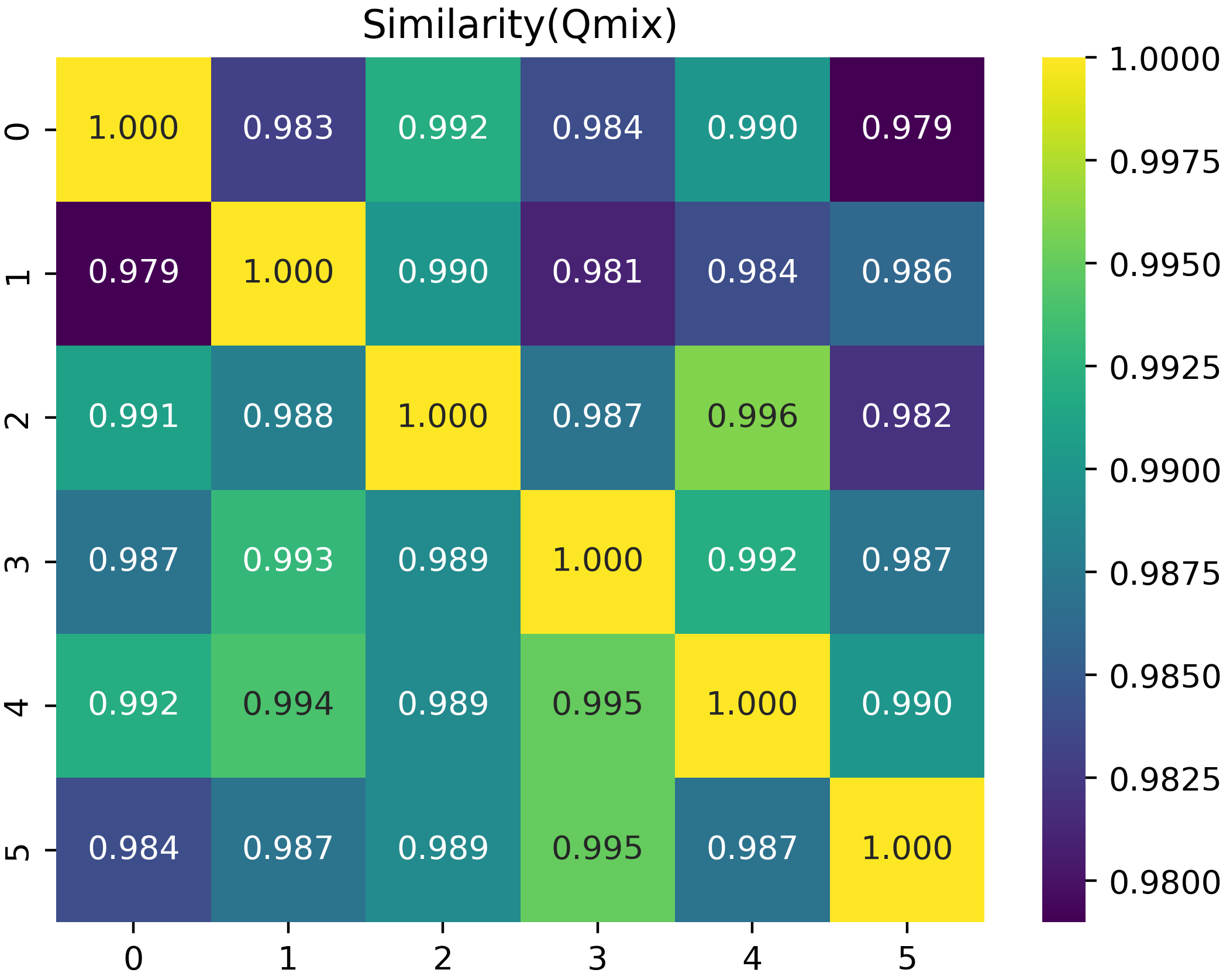}
    \end{minipage}
    }
    \caption{\textbf{Graphical illustration of agent diversity}. (a) shows how the agent within the red box should operate, with Q-values for actions. We calculated the Q-values of different agents and displayed three representative ones. For the same observation, Agent 1 moves southward to protect teammates; Agent 3 attacks low-health enemies; Agent 6 moves to the upper left to pull enemies away. We calculated the cosine similarity between the Q-values of other agents and the current agent's Q-values. (b) and (c) compare our method with Qmix, showing cosine similarity with other agents under the same observation. Our agents exhibit more diversity.}

    \label{similarity}
\end{figure*}

Due to the utilization of parameter-sharing techniques in most networks, such as QMIX, updating one agent triggers updates across all agent networks, which evidently does not align with our requirements. What we desire is diversity and specialization among each agent, so that even under the same observations, they exhibit distinct behaviors. However, if we allow parameters to differ among agents, both the parameter count and training difficulty would increase. To address this issue, inspired by the CDS algorithm\cite{li2021celebrating}, we partition the critic network into shared and independent layers, and the $Q$ value of agent $i$ executing  $a_i$ in observation $o_i$ is:
\begin{equation}
    Q_\text{tot}^i(o_i,a_i) = Q_\text{com}(o_i,a_i)+\lambda Q_\text{sep}^i(o_i,a_i).
\end{equation}
Here, $\lambda$ controls the scale of separate $Q$ values. The shared layer facilitates parameter sharing among all agents, aiding them in extracting overall features from the environment, which is similar to QMIX. This helps establish a consensus among agents regarding the value of observations and states, for instance, recognizing that attacking teammates is undesirable while attacking enemies is valuable. On the other hand, the independent layer embodies the diversity among agents. Taking a two-agent collaborative game as an example, some agents may play the role of ``tank'', where executing aggressive actions in the current state is highly beneficial. Conversely, other agents playing the role of ``attacker'' may find aggressive actions unproductive, as there are already agents absorbing enemy fire, and thus opt not to attack. Hence, the independent layer serves to showcase individuality and roles among agents, with the final output of the critic network obtained through a weighted sum of the shared and independent layers.

In the process of performing additional training on agents, to ensure that only the current agent is updated without affecting others, we refrain from computing gradients for the shared layer during parameter updates and only update the network parameters of each agent's independent layer. This approach not only avoids interdependence among agents but also speeds up the algorithm by performing partial gradient backpropagation. Only during overall data updates are both the shared and independent layers simultaneously updated across all agents.

\subsection{Update of the MANGER Agents}

The updated formulas for the total Q-value in QMIX and parameters are as follows:
\begin{equation}
\small Q_\text{tot}(s,a;\theta) = f_\phi(Q^1(o_1,a_1;\theta_1),...,Q^N(o_N,a_N;\theta_N)),
\end{equation}

\begin{equation}
 \small y = r(s,a)+\gamma \times \max_{a^{'}} Q_\text{tot}(s',a';\theta),
\end{equation}

\begin{equation}
\small \theta \gets \theta  + \alpha \times (Q_\text{tot}(s,a;\theta)-y).
\end{equation}
Here, $f$ denotes the mixer network and $\alpha$ is the learning rate. By updating the parameters of the mixer network and the corresponding Q networks of each agent, the agents can estimate the overall Q value more accurately.

Comparing to QMIX, the update equations for MANGER agents are as follows:

\begin{equation}
\small Q_\text{tot}(s,a;\theta) = f_\phi(Q^1_\text{tot}(o_1,a_1;\theta_1),...,Q^N_\text{tot}(o_N,a_N;\theta_N)),
\end{equation}

\begin{equation}
 \small y = r(s,a)+\gamma \times \max_{a^{'}} Q_\text{tot}(s',a';\theta),
\end{equation}

\begin{equation}
\small \theta \gets \theta  + \alpha \times (Q_\text{tot}(s,a;\theta)-y) \times {H}({T}),
\end{equation}

\begin{equation}
{T} \gets {T} - {1}.
\end{equation}

Here, ${H}({x})=[h(x_1),h(x_2),...,h(x_N)]$ , $h(x)=\left\{\begin{array}{ll}
1, & x>0 \\
0, & x \leq 0
\end{array}\right.$ and ${T} = [T_1,T_2,...,T_N]$ where $T_i$ is calculated by equations \eqref{eq1} and \eqref{eq2}. The formula described above will cyclically compute and continuously update the parameters until all components of ${T}$ are not greater than 0. When the novelty corresponding to agent $i$ is low, $h(T_i)$ will be 0 and will be ignored during parameter updates; when the degree of novelty is high, $T_i$ will be greater than 1 and $h(T_i)$ will equal 1, which is equivalent to an additional update to the parameters $\theta_i$ corresponding to $Q_i$.

\section{Experiment}

In this section, we will experimentally address the following questions: (1) Can our method improve sample efficiency, leading to higher or faster convergence of agents' win rates across different tasks? (2) Do agents exhibit diverse behaviors through updates of varying frequencies? (3) Is our proposed RND standalone update module effective compared to the holistic update? (4) Is our algorithm's improvement due to adding the $Q_{sep}$ module? To answer these four questions, we conducted the following experiments for validation.

\begin{figure*}[t]
    \centering
    \begin{minipage}[t]{0.46\linewidth}
    \centering
    \includegraphics[width=1.0\linewidth,height=0.65\linewidth]{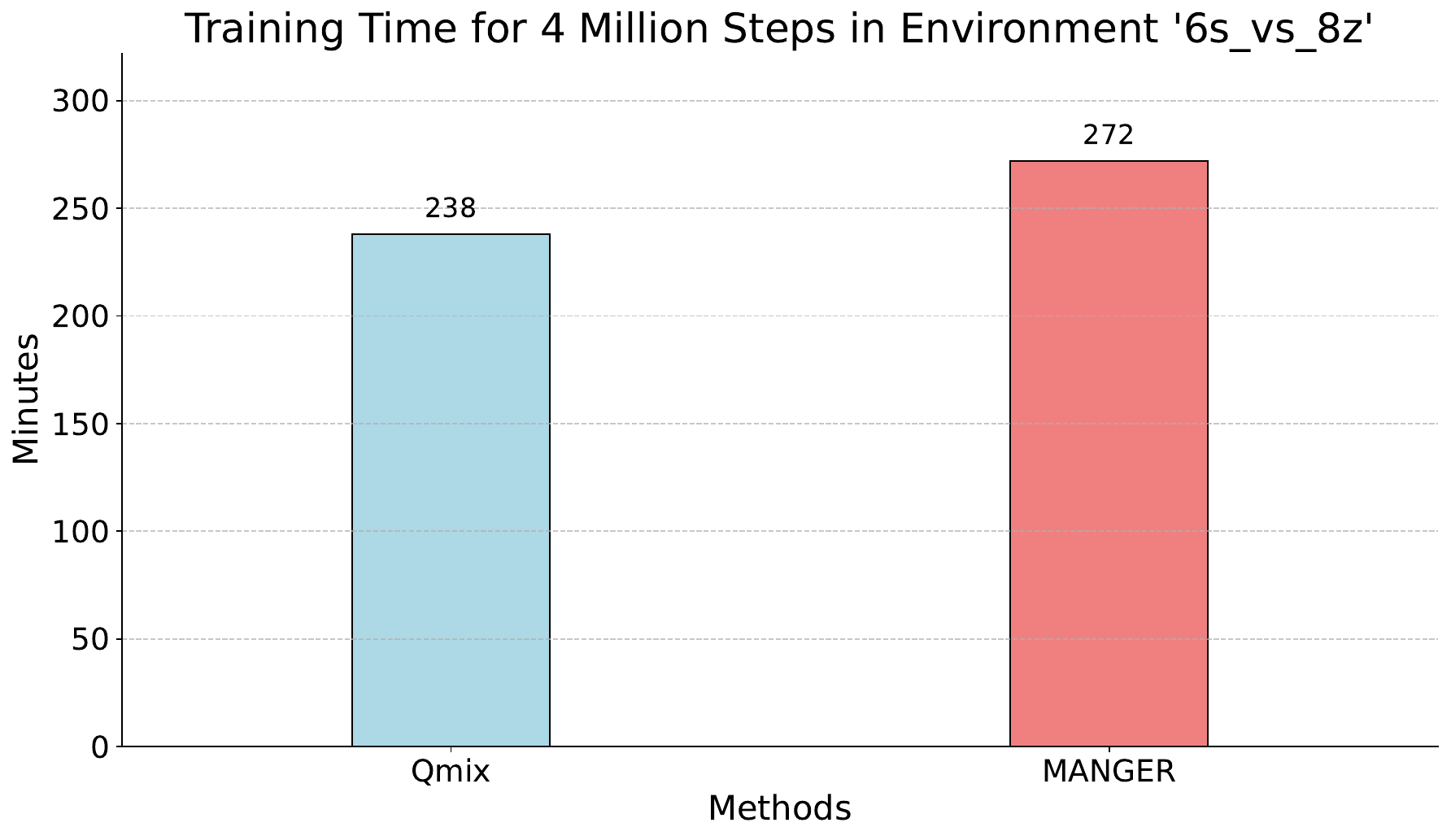}
    \end{minipage}
    \begin{minipage}[t]{0.46\linewidth}
    \centering
    \includegraphics[width=0.8\linewidth,height=0.65\linewidth]{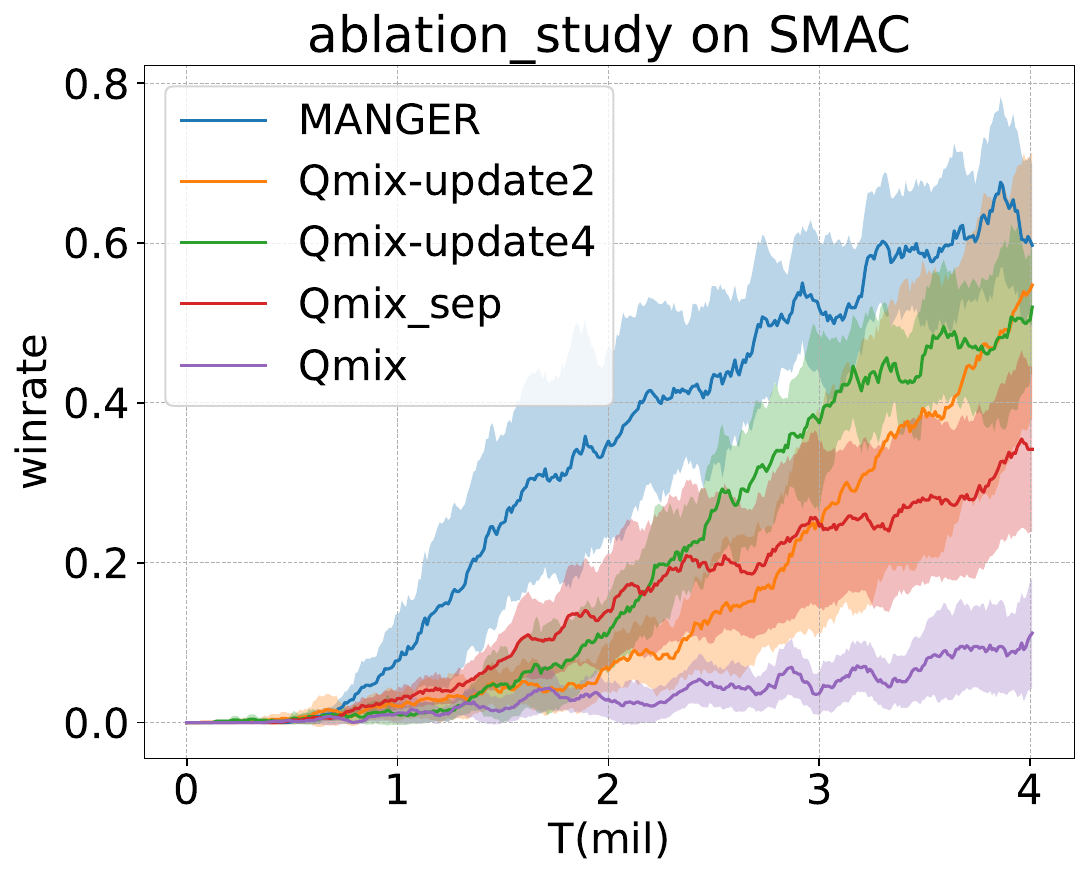}
    \end{minipage}
    \caption{\textbf{Left:} Training time comparison between the MANGER and Qmix methods in the 6h\_vs\_8z environment. \textbf{Right:} Ablation study of MANGER in the 6h\_vs\_8z environment. It can be observed that our method does not significantly increase training time while achieving performance improvements with the same number of environment interactions. The ablation study further confirms that the performance gains are not solely due to increased sample reuse or network decomposition but are specifically attributed to the targeted utilization of certain important samples by the RND module.}
    \label{ablation}
\end{figure*}

\textbf{Setup} To validate the effectiveness of our proposed algorithm, we employed the widely used StarCraft Multi-Agent Challenge (SMAC, \cite{samvelyan2019starcraft}) in multi-agent reinforcement learning. SMAC includes various environments that require cooperation and coordination and is based on the popular real-time strategy game StarCraft II, providing a diverse set of tasks for multi-agent collaboration. We also used the Google Research Football (GRF) \cite{kurach2020google} environment, which contains numerous multi-agent tasks where agents must cooperate to achieve goals, serving as a testbed for assessing the effectiveness of algorithms in handling complex, real-world-like multi-agent scenarios. We utilized PyMARL2 \cite{hu2021rethinking} as our codebase and employed QMIX as our baseline algorithm, evaluating whether QMIX combined with our method yields superior performance. We compared our approach against several popular methods, including QMIX, QPLEX, and Qatten, using the parameters recommended in the respective papers. 

\textbf{Result} In this section, we will address the question (1) and validate whether our algorithm can enhance the convergence rate of agents, thus accelerating the improvement of win rates. Within the SMAC environment, we have chosen a variety of tasks to evaluate the effectiveness of our algorithm, including both symmetric and asymmetric environments. We recorded the test win rates of each method on various tasks and compared the final performance and convergence rates of different methods. We plotted win rate curves of different methods under various task environments for comparison, as shown in Figure \ref{fig:exp_on_smac}. It can be observed that in the majority of environments, our method is able to significantly enhance the performance of the baseline QMIX algorithm within a short period. In the GRF environment, we selected three more challenging settings, and the experimental results are shown in Figure 2. It can be seen that our method performs the best among all the methods. By leveraging sample reuse, our method enables agents to better utilize data, thereby improving task win rates, thus validating the effectiveness of our algorithm. Furthermore, compared to other updated improvement algorithms like Qatten and QPLEX, our method achieves the best performance across all methods, further demonstrating its effectiveness. 

\textbf{Analysis} To answer the question (2), whether agents have demonstrated differentiated performance, we conducted both qualitative and quantitative analyses, as shown in Figure \ref{similarity}. During the visualization evaluation, we selected a representative frame to observe the Q values of different agents under the same observation and their corresponding strategies. It can be seen that under the same observation, the Q values of different agents trained by MANGER have significant differences, and the actions they take are also different. For the QMIX algorithm, this would be impossible, as the Q values would be very close under the same observation. Through a quantitative comparison using heat maps, it can be seen that the Q value similarities among different agents trained by our method are lower, showing more differentiation and division of labor, which proves the effectiveness of our algorithm. In addition, we have also visualized the environment to observe whether our algorithm can enable agents to achieve division of labor and generate diversity, as shown in Figure \ref{visualize}. It can be seen that agents can take on the role of a tank that actively absorbs damage to help teammates with output; activate the offensive role of attackers; attract hatred to pull monsters, avoiding the continuous activation of the tank by roaming characters.

\textbf{Ablation Study} In this section, we will verify questions (3) and (4). To validate the effectiveness of individually updating each datum with the RND module, we conducted statistical analysis on the average individual update counts. We found that due to the nature of the normal distribution, the extra average update times tend to be about 0.5, which does not significantly increase the training cost. We conducted comparisons in some environments, and the comparison of training time and the final results are shown in Figure \ref{ablation}. It can be observed that compared to QMIX, QMIX with an added separate value, and methods that fix multiple updates on this basis, our method achieves better training results with relatively fewer updates. This demonstrates that utilizing the RND network for individual updates enhances effectiveness not merely by increasing the average additional update counts, but rather by targeting additional updates towards crucial states and trajectories. Additionally, the performance of adding a separate module to QMIX not being as good as MANGER also proves that the previous results were not due to the introduction of new modules, but rather the outcome of repeated updates by the RND module.

\section{Conclusion}
To improve sample efficiency in multi-agent systems and enhance performance diversity among different agents, enabling them to develop specialization for collaborative task completion, this paper proposes the MANGER method to address this issue. By using the RND method to calculate the novelty of each agent's observations, states with high novelty and inaccurate Q-value estimates are updated multiple times, while states with low novelty are not given additional updates. This approach enables differentiated updates for different agents, increases sample efficiency, and allows for targeted extra updates on rare samples. Experimental results show that our method, which only provides additional updates to novel states, does not introduce significant additional time overhead. In SMAC and GRF environments, our method outperforms QMIX, Qatten, and Qplex algorithms, achieving higher win rates within fewer training steps. By observing the similarity of Q-values among different agents and through visualization, we demonstrate that MANGER can enhance the diversity of agent policies and enable specialization for collaborative task completion, proving the effectiveness and feasibility of our method.

\newpage
\section{Acknowledgements}

This work was supported by the STI 2030-Major Projects under Grant 2021ZD0201404. The authors also thank the anonymous reviewers for valuable comments.

\bibliography{aaai25}
\onecolumn 
\newpage
\clearpage
\section{A. MANGER  Pseudo-code}

In this section, we present the detailed pseudo-codes for MANGER.

\begin{algorithm}
\caption{MANGER}
\begin{algorithmic}[1]
\STATE \textbf{Require:} Number of training interval $M$, RND network update interval $M_{\mathrm{rnd}}$, coefficient for $Q_i$ network $\lambda$, and coefficient for additional updates $\alpha$.
\STATE Initialize parameters for common Q-networks $\theta_{\mathrm{com}}$, separate Q-networks $\theta_{\mathrm{sep}}^{1,2,...,j}$, mixing network $\phi$ and RND predict network $\zeta$
\STATE Initialize target networks $\theta_{\mathrm{com}}^{'}$, $\theta_{\mathrm{sep}}^{'1,2,...,j}$, $\phi^{'}$ and $\zeta^{'}$
\STATE Initialize experience replay buffer $D$

\FOR{each episode}
    \STATE Reset the environment and get initial state $s$
    \FOR{each time step $S$ in episode}
        \FOR{each agent $i$}
            \STATE Calculate $Q_{\mathrm{tot}}^{i}$ = $Q_{\mathrm{com}}^{i}$ + $\lambda Q_{\mathrm{sep}}^{i}$
            \STATE Select action $a_i$ using $\epsilon$-greedy policy based on $Q_{\mathrm{tot}}^{i}(s, a_i; \theta_{\mathrm{commom}}, \theta_{\mathrm{sep}}^{i})$
        \ENDFOR
        \STATE Execute joint action $a = \{a_1, \dots, a_n\}$, observe reward $r$, new state $s'$
        \STATE Store transition $(s, a, r, s')$ in buffer $D$

        \IF{mod($S$, $M$) == 0}
            \STATE Sample a batch of transitions from $D$
            \STATE Calculate target Q-value: $y = r + \gamma \cdot \max Q(s', a'; \theta')$
            \STATE Calculate $N_{\mathrm{i}}$ for each agent using Equation \ref{eq1}
            \STATE Calculate extra updating times $T_{\mathrm{i}}$for each agent using Equation \ref{eq2}
            \STATE Calculate local Q-values for each agent $Q_i(s, a_i; \theta)$
            \STATE Calculate global Q-value $Q_{total} = MixingNetwork(\{Q_1, \dots, Q_n\}; \phi)$
            \STATE Compute loss: $L = (Q_{total} - y)^2$
            \IF{mod($S$, $M_{\mathrm{rnd}}$) == 0}
                \STATE Compute RND loss: $L_{\mathrm{rnd}} = \|f_\text{target}(o_i)-f_\text{predictor}(o_i)\|^2$
                \STATE Update $\zeta$ using gradients of $L_{\mathrm{rnd}}$
            \ENDIF
            \STATE Update $\theta_{\mathrm{com}}$, $\theta_{\mathrm{sep}}^{'1,2,...,j}$ and $\phi$ using gradients of $L$
            \FOR{$T_{\mathrm{1,2,...j}}$}
                \STATE Compute loss:  $L_{extra} = (Q_{extra} - y)^2$
                \STATE Update $\theta_{\mathrm{sep}}^{'1,2,...,j}$ and $\phi$ using gradients of $L_{\mathrm{extra}}$
            \ENDFOR
            \STATE Update target networks $\theta' = \tau\theta + (1-\tau)\theta'$ and $\phi' = \tau\phi + (1-\tau)\phi'$
        \ENDIF
    \ENDFOR
\ENDFOR

\end{algorithmic}
\end{algorithm}

\section{B. Experimental Settings}

The \texttt{Q-Value} neural network structure calculating for agent Q value consists of a linear layer that transforms the input into a hidden representation, followed by a GRU cell that processes temporal dependencies. After the GRU, the output is passed through another linear layer to compute the main \(Q\)-values for each action. Additionally, the network incorporates a set of multi-layer perceptrons (MLPs), where each agent in the system is assigned one MLP consisting of a single linear layer. These MLPs take the GRU output and generate agent-specific \(Q_{\text{sep}}\) values. The final \(Q\)-values, \(Q_{\text{sum}}\), are computed by combining the main \(Q\)-values and the agent-specific \(Q_{\text{sep}}\) values. We use ReLU as our activate function.

The \texttt{QMixer} network integrates individual agent \(Q\)-values into a global \(Q_{\text{tot}}\) value using several layers. Initially, hypernetworks generate weights \(w_1\) and \(w_{\text{final}}\) based on the global state. The agent \(Q\)-values are mixed through a hidden layer with ELU activation. A state-dependent bias is added, and the final global \(Q_{\text{tot}}\) value is computed by applying \(w_{\text{final}}\) and combining with a state-dependent value function \(V(s)\). The \texttt{RND} network consists of two components: the target network and the predictor network. Both networks are composed of 2 linear layers with ReLU activation functions. The target network's parameters are fixed and do not require gradients, while the predictor network's parameters are updated during training. The RND network calculates observation novelty based on the squared difference between the outputs of the target and predictor networks.


Our experiments were performed by using the following hardware and software on SMAC env:

\begin{itemize}
    \item GPUs: NVIDIA GeForce RTX 3090
    \item CPU: AMD EPYC 7282 16-Core Processor
    \item Python: 3.7.16
    \item CUDA: 11.6
    \item numpy: 1.21.6
    \item pytorch: 1.13.1
    \item gymnasium: 0.28.1
    \item gym: 0.11.0
    \item pygame: 2.1.0
    \item protobuf: 3.20.0
    \item pysc2: 3.0.0
    \item smac: 1.0.0
    
\end{itemize}

And our experiments conducted on GRF were performed by the following hardware and software:

\begin{itemize}
    \item GPU: NVIDIA GeForce RTX 3070 Ti
    \item CPU: 12th Gen Intel(R) Core(TM) i9-12900H
    \item Python: 3.8.10
    \item CUDA: 12.3
    \item numpy: 1.24.4
    \item pytorch: 2.4.0+cu121
    \item gym: 0.25.2
    \item pygame: 2.1.2
    \item gfootball: 2.10.2
\end{itemize}

\section{C. Hyperparameters}

The hyperparameter settings we use for the SMAC and GRF environment are as follows. For hyperparameters common across different algorithms, we set the same values to ensure comparability. Between different environments, we refer to the parameters provided by the original framework and make adjustments to achieve optimal performance.

\begin{table}[h!]
\centering
\resizebox{0.8\columnwidth}{!}{
\begin{tabular}{ccc}
\toprule
\textbf{Name} & \textbf{Description} & \textbf{Value} \\
\midrule
\textit{lr} & learning rate for Q-network & 1e-3 \\
\textit{lr\(_{\text{rnd}}\)} & learning rate for RND-network & 1e-3 \\
\textit{optimizer} & type of optimizer & Adam \\
\textit{batch\_size} & batch\_size for training & 128 \\
\textit{batch\_size\_run} & parallel env collecting data & 8 \\
\textit{buffer\_size} & buffer\_size for training & 5000 \\
\textit{mixing\_embed\_dim} & embed\_dim for mixing network & 32 \\
\textit{$\gamma$} & reward\_decay\_factor & 0.99 \\
\textit{$M$} & total steps for training & 4000000 \\
\textit{$M_{\mathrm{target}}$} & interval step for update target network & 200 \\
\textit{$M_{\mathrm{rnd}}$} & interval step for RND network & 2 \\
\textit{$M_{\mathrm{anneal}}$} & choose action randomness decay time & 100000 (500000 for \textit{6h\_vs\_8z}) \\
\textit{$\epsilon_{\mathrm{start}}$} & start prob for choosing random actions & 1.0 \\
\textit{$\epsilon_{\mathrm{finish}}$} & final prob for choosing random actions & 0.05 \\
\textit{$TD_{\mathrm{\lambda}}$} & ratio to control Q-value update & 0.6 (0.3 for \textit{6h\_vs\_8z}) \\
\textit{$\alpha$} & coefficient for additional updates & 1 \\
\textit{$\beta$} & max number additional updates & 3 \\
\textit{$\lambda$} & coefficient for $Q_i$ network & 0.5 \\
\bottomrule
\end{tabular}
}
\caption{Hyperparameters on SMAC envs}
\label{tab:SMAC env}
\end{table}

\begin{table}[h!]
\centering
\resizebox{0.6\columnwidth}{!}{
\begin{tabular}{ccc}
\toprule
\textbf{Name} & \textbf{Description} & \textbf{Value} \\
\midrule
\textit{lr} & learning rate for Q-network & 5e-4 \\
\textit{lr\(_{\text{rnd}}\)} & learning rate for RND-network & 5e-4 \\
\textit{optimizer} & type of optimizer & Adam \\
\textit{batch\_size} & batch\_size for training & 128 \\
\textit{batch\_size\_run} & parallel env collecting data & 16 \\
\textit{buffer\_size} & buffer\_size for training & 2000 \\
\textit{mixing\_embed\_dim} & embed\_dim for mixing network & 32 \\
\textit{$\gamma$} & reward\_decay\_factor & 0.999 \\
\textit{$M_{\mathrm{target}}$} & interval step for update target network & 200 \\
\textit{$M_{\mathrm{rnd}}$} & interval step for RND network & 2 \\
\textit{total\_training\_step} & total steps for training & 4000000 \\
\textit{epsilon\_anneal\_time} & choose action randomness decay time & 500000 \\
\textit{epsilon\_start} & start prob for choosing random actions & 1.0 \\
\textit{epsilon\_finish} & final prob for choosing random actions & 0.05 \\
\textit{TD\(_{\text{lambda}}\)} & ratio to control Q-value update & 1.0 \\
\textit{$\alpha$} & coefficient for additional updates & 1 \\
\textit{$\beta$} & max number additional updates & 3 \\
\textit{$\lambda$} & coefficient for $Q_i$ network & 0.5 \\
\bottomrule
\end{tabular}
}
\caption{Hyperparameters on GRF envs}
\label{tab:GRF env}
\end{table}

\section{D. More Results}
\begin{figure*}[h]
    \centering
    \includegraphics[width=0.38\linewidth,height=0.225\linewidth]{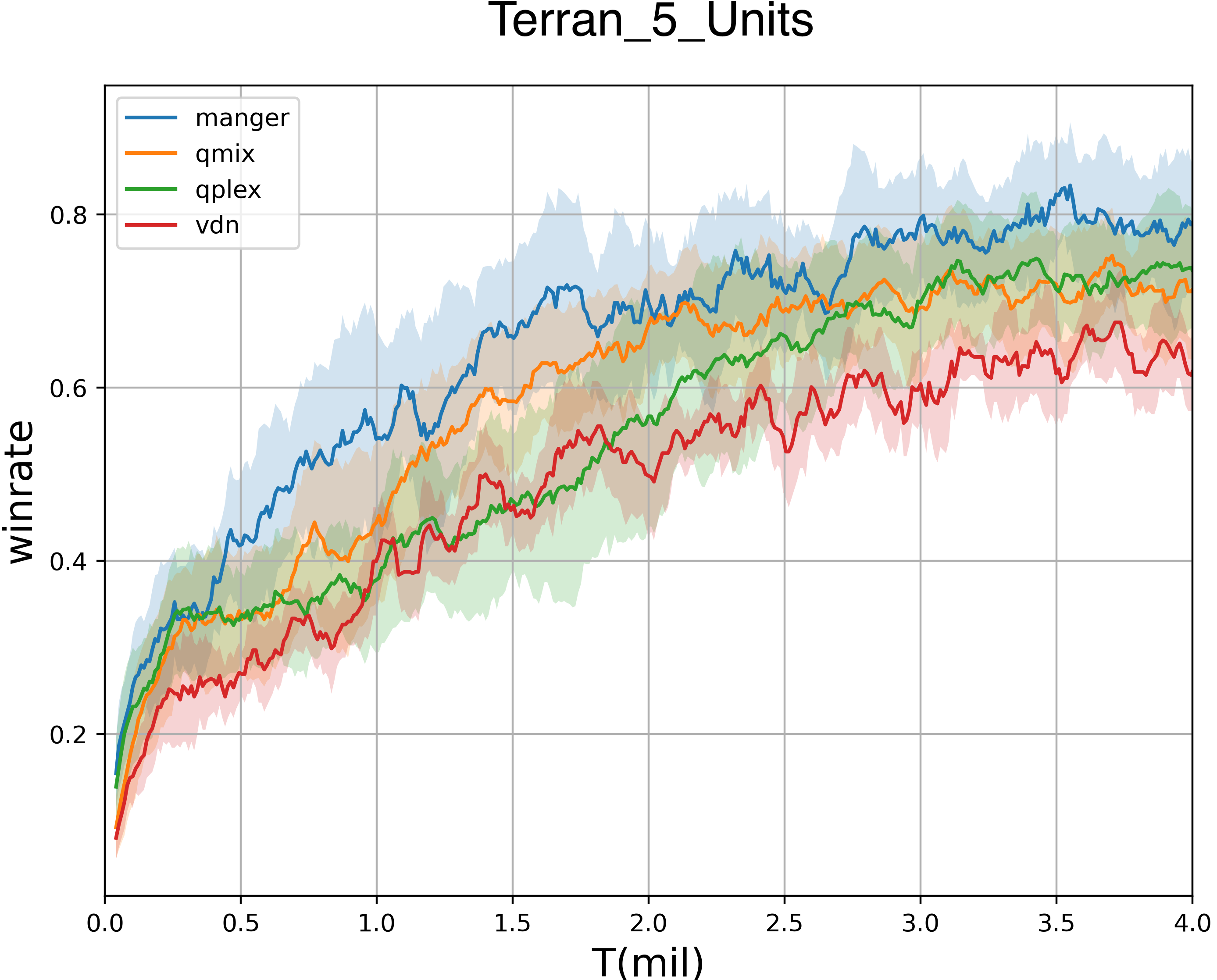 }
    
    \includegraphics[width=0.35\linewidth,height=0.23\linewidth]{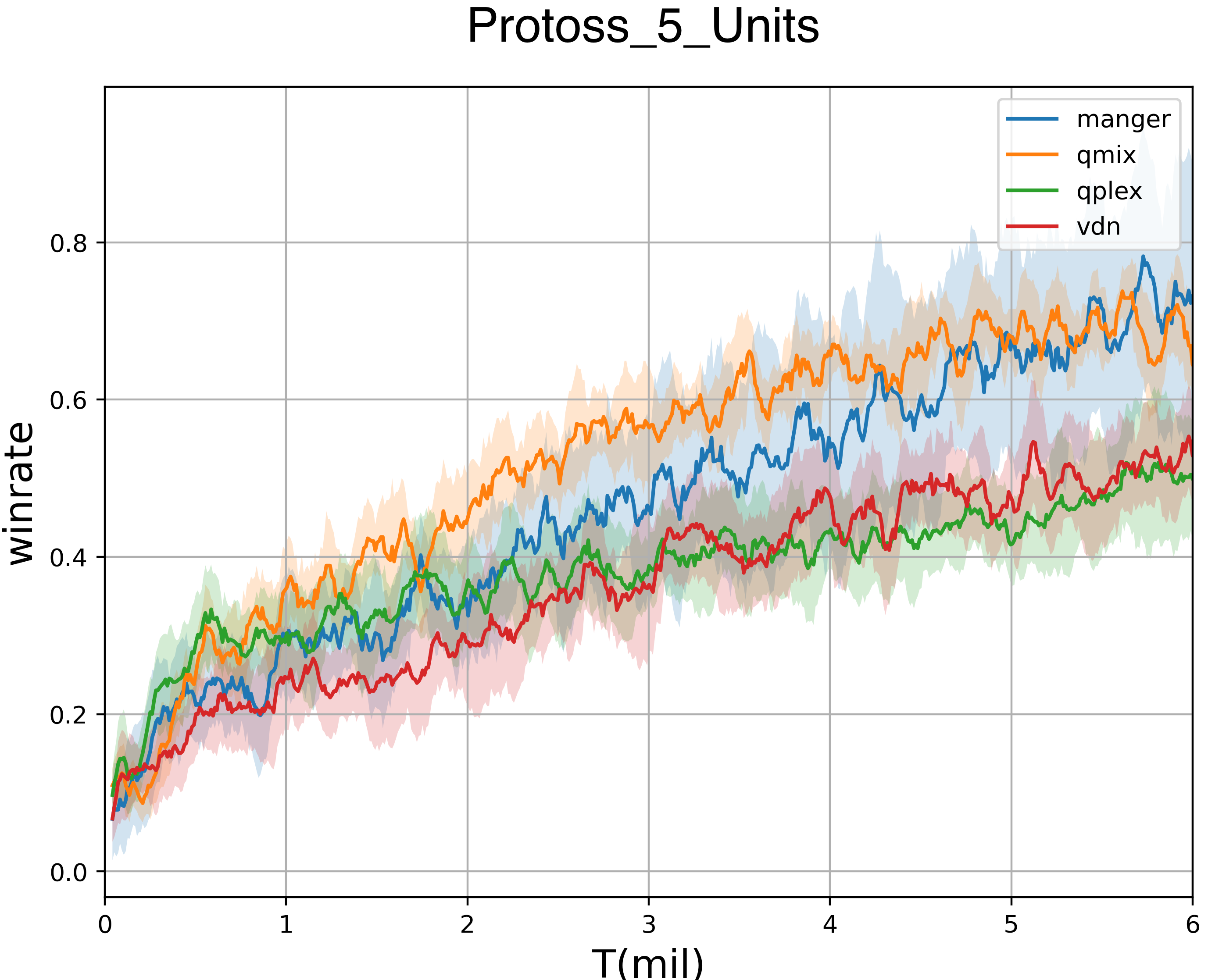}
    \includegraphics[width=0.34\linewidth,height=0.23\linewidth]{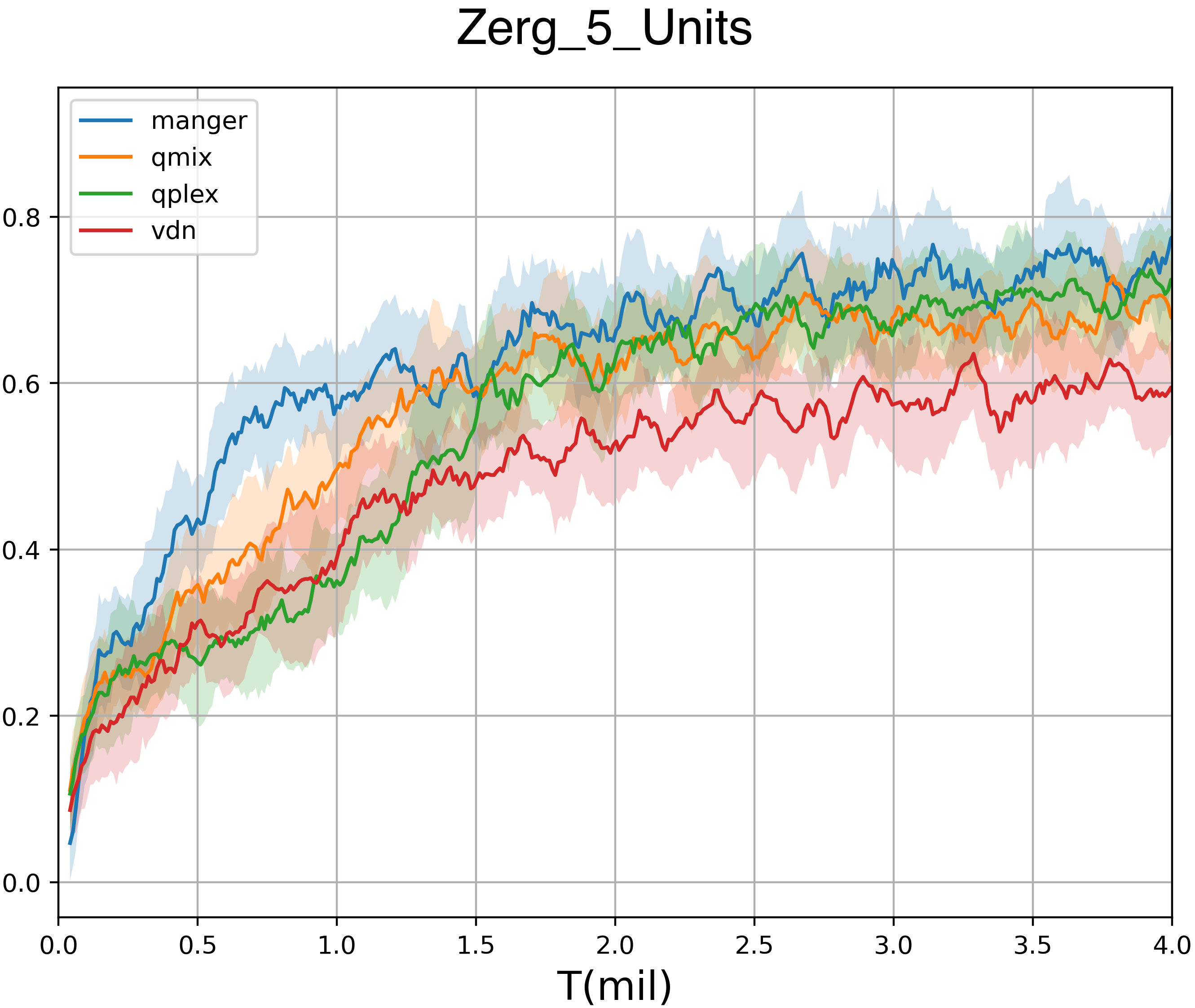}
    \caption{\textbf{Experiments on SMAC-V2 environments}. All curves are averaged over 5 independent runs.}
    \label{fig:smacv2}
\end{figure*}

To further verify the effectiveness of the MANGER method, we conducted experiments in the latest SMAC-V2 \cite{ellis2024smacv2} environment. We used the pymarl3 \footnote{https://github.com/tjuHaoXiaotian/pymarl3} library as our codebase, and the baseline algorithms were implemented with their default recommended parameters. The experimental results are shown in Figure \ref{fig:smacv2}. As can be seen, in three tasks, our algorithm generally achieved the best performance, further demonstrating that MANGER performs well even in challenging environments. This validates the effectiveness and robustness of the method.

\end{document}